\documentclass{article}
\usepackage{spconf,amsmath,graphicx,url,bbm,comment,amsfonts}
\usepackage[caption=false]{subfig}
\usepackage{color}
\usepackage{eso-pic}

\title{Bit Allocation for Multi-Task Collaborative Intelligence}
\name{{Saeed Ranjbar Alvar and Ivan V. Baji\'c}}
\address{School of Engineering Science, Simon Fraser University, Burnaby, BC, Canada 
}
\begin{document}
%
 \AddToShipoutPicture*{\small \sffamily\raisebox{1.0cm}{\hspace{4.4cm}Copyright 
\copyright \hspace{1mm} 2020 IEEE. The original publication is available for download at ieeexplore.ieee.org.
}}

\maketitle
\begin{abstract}
Recent studies have shown that collaborative intelligence (CI) is a promising framework for deployment of Artificial Intelligence (AI)-based services on mobile devices. In CI, a deep neural network is split between the mobile device and the cloud. Deep features obtained at the mobile are compressed and transferred to the cloud to complete the inference. So far, the methods in the literature focused on transferring a single deep feature tensor from the mobile to the cloud. Such methods are not applicable to some recent, high-performance networks with multiple branches and skip connections. In this paper, we propose the first bit allocation method for multi-stream, multi-task CI. We first establish a model for the joint distortion of the multiple tasks as a function of the bit rates assigned to different deep feature tensors. Then, using the proposed model, we solve the rate-distortion optimization problem under a total rate constraint to obtain the best rate allocation among the tensors to be transferred. Experimental results illustrate the efficacy of the proposed scheme compared to several alternative bit allocation methods.  
\end{abstract}
%


\section{Introduction}
\label{sec:intro}
Recent mobile devices are increasingly capable of running artificial intelligence (AI) based applications~\cite{ai_benchmark}. However, due to the limitations on battery and processing power of these devices, the most sophisticated AI models, built around large, deep neural networks (DNNs), can still only run in the cloud. The current practice is to send the data from the mobile to the cloud for AI-based processing, and then receive the results back on the mobile. However, recent studies~\cite{neurosurgeon,Eshratifar_offloading_2018} have shown that in many cases, a more efficient approach from an energy and latency point of view, is to perform some computations on the mobile and send the results of those computations (rather than the original data) for further processing in the cloud. This approach has been termed \emph{collaborative intelligence} (CI). In CI, usually, the initial layers of a DNN are deployed on the mobile and the remainder of the model is in the cloud. The mobile sends a tensor of deep features to the cloud for further processing.  


Since the deep feature tensor needs to be transmitted to the cloud, the tensor should be compressed prior to transmission in order to utilize the communication channel efficiently. Several papers have studied various ways to compress such tensors.
In~\cite{choi_icip,choi_lossless,battlefiled}, the authors proposed lossy and lossless methods for encoding the deep feature tensor. The authors of~\cite{bottlenet,butterfly} introduced dimension reduction units in order to reduce the number of features for the purpose of compression. In~\cite{saeed_icip}, a loss function encouraging feature compressibility has been used as a part of the overall training loss for multi-task learning, resulting in more compressible features. 

All the studies mentioned above focused on compressing one feature tensor, sometimes referred to as ``bottleneck features,'' taken from one layer in the DNN. This is appropriate for single-stream networks such as VGG~\cite{VGG} and its many derivatives. However, recent high-performance DNN architectures based on residual blocks~\cite{resnet} and dense blocks~\cite{densenet} often involve multiple streams of features through various skip connections. In such cases, the mobile may need to transmit multiple feature tensors to the cloud. Hence, a natural question arises - how do we allocate bits among multiple feature tensors to maximize the model's accuracy? This is the main focus of the present paper.  

It should be noted that bit allocation for DNN compression has recently been studied in~\cite{Girod_ICIP2019}, where the authors propose a strategy for allocating bits to both weights and activations of a DNN, with a focus on single-stream, single-task DNNs. The present paper differs from~\cite{Girod_ICIP2019} in several ways: (1) our focus is on multi-task, multi-stream DNN; (2) we focus on bit allocation among multiple feature tensors (i.e., activations), while the weights of the original DNN are unchanged; (3) we provide a convex approximation to the model's rate-distortion surface, which allows a closed-form solution of the bit allocation problem, unlike~\cite{Girod_ICIP2019}, where the solution is found by search. 


The paper is organized as follows. Section~\ref{sec:proposed} presents the formulation of the bit allocation problem among multiple feature tensors, followed by a convex approximation to the rate-distortion surface, and a closed-form solution. Section~\ref{sec:Experiments} presents experimental results that illustrate the efficacy of the derived solution, followed by conclusions in Section~\ref{sec:conclusion}. 

\begin{figure*}[t!]	
	\centering
	\centerline{\includegraphics[width=\textwidth]{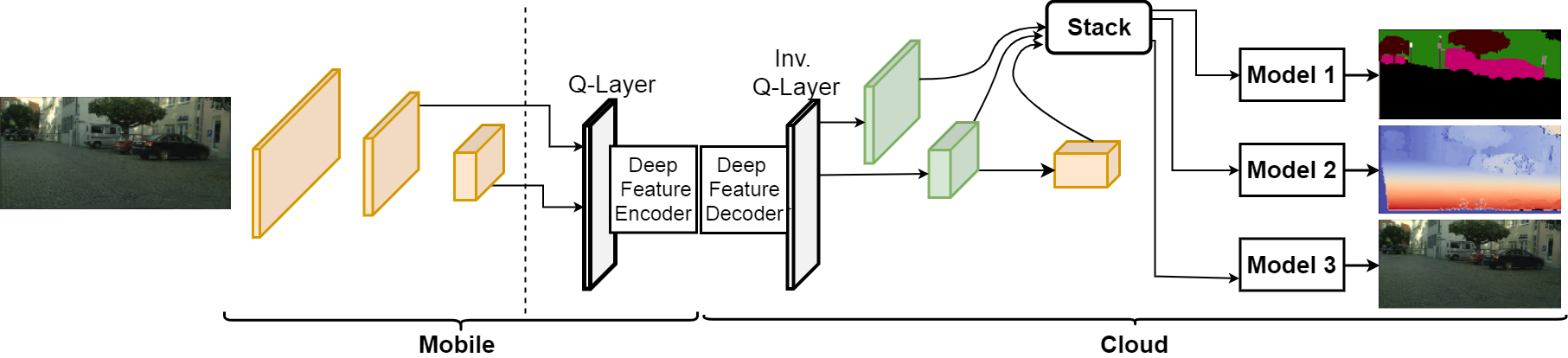}}
	\caption{A two-stream, three-task DNN used in our experiments. For clarity, the illustration focuses on the feature tensors, rather than layers. The vertical dashed line shows where the DNN streams are cut, and the two arrows indicate the two feature streams (tensors) that will be 
	quantized, compressed, and then transferred to the cloud. Computed tensors are shown in yellow, while decoded tensors are shown in green. 
	} 
	\label{fig:model}
\end{figure*}

\section{Proposed Method}
\label{sec:proposed}

Fig.~\ref{fig:model} shows a two-stream, three-task DNN that will be used in our experiments. This DNN is shown for illustration purposes, and to ground the discussion around experiments in Section~\ref{sec:Experiments}. The methodology presented below, however, applies to more general multi-stream, multi-task DNNs.  

\subsection{Joint distortion over multiple tasks}
\label{subsec:distortion}
Consider a DNN model with $M$ tasks. Bit allocation for a multi-task model can be considered a multi-objective optimization. A popular way to deal with such problems is to \emph{scalarize} them by defining a single cost function that captures all the individual objectives~\cite{Chong_Zak_Optim_2013}. In our case, we will define multi-task distortion that will be a positive linear combination of distortions of individual tasks, as described below. 

The trained multi-task model is initially evaluated without feature tensor compression.  
This evaluation leads to the best possible performance on each task. Let $\overline{A_i}$ be the the model's average performance on the $i$-th task, on a given dataset, without tensor compression. We define the task-specific distortion as the fraction of the performance drop relative to the case where no compression is applied to the feature tensors. Let $A_i$ be the average performance with tensor compression on the same dataset. Then the distortion for task $i$ is defined as
\begin{equation}
\label{eq:dist_def}
    D_i = \frac{\overline{A_i} - A_i}{\overline{A_i}}.
\end{equation}
The total multi-task distortion is the positive linear combination of  task-specific distortions,
\begin{equation}
\label{eq:total_dist_def}
    D_t = \sum_{i=1}^{M} w_i D_i, 
\end{equation}
where $w_i>0$ is the weight corresponding the $i$-th task. Since the relative importance of tasks may vary depending on the application, $w_i$'s can be tuned as needed. In the experiments, we will explore several settings for $w_i$'s. 

\subsection{Bit allocation among multiple deep feature tensors}
\label{subsec:bit_allocation}
\begin{figure}[t]	
	\centering
	\centerline{\includegraphics[scale=0.31]{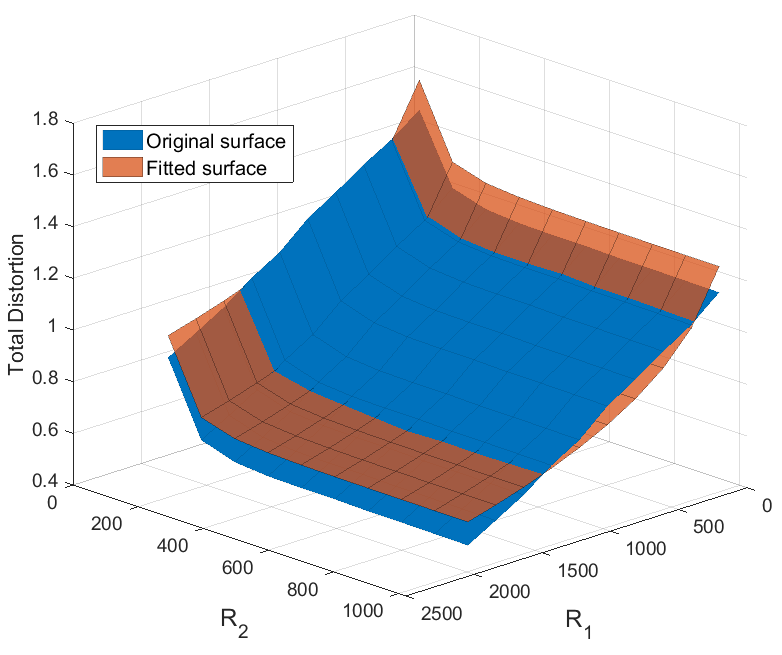}}
	\caption{Rate-distortion surface obtained by encoding 
	two deep feature tensors (blue) and the fitted surface (orange). 
	$R_1$ and $R_2$ are the average bit rates (kbits/tensor) of the two tensors.} 
	\label{fig:surface}
\end{figure}

Let $\{\mathbf{X}_1,\mathbf{X}_2,...,\mathbf{X}_N\}$ be the deep feature tensors to be compressed, and let  $\{R_1,R_2,...,R_N\}$ be the corresponding bit rates of the encoded tensors. All task-specific distortions~(\ref{eq:dist_def}), as well as the total multi-task distortion~(\ref{eq:total_dist_def}), are functions of the rates $R_j$. 
The goal is to select $R_j$ so as to minimize the total multi-task distortion~(\ref{eq:total_dist_def}), subject to the total rate constraint:
\begin{equation}
\begin{aligned}
\label{eq:rdo}
&\operatorname*{arg\min}_{(R_1,...,R_N)\in \mathbb{R}^N} D_t(R_1,..., R_N) \\
&\enspace \text{ s. t.} \quad  \sum_{j=1}^{N} R_j \leq R_t 
\end{aligned}
\end{equation}
In order to solve~(\ref{eq:rdo}), we proceed in the following way. We measure $D_t(R_1,...,R_N)$ on a set of rate tuples $(R_1,...,R_N)\in \mathcal{R}$ (the vector of feasible rates for tensors), on a given dataset, and then fit a convex surface to the measured distortions points. The benefit of this approach is that, with $D_t(R_1,...,R_N)$ convex, problem~(\ref{eq:rdo}) has a closed-form solution, hence bit allocation can be computed easily even if $R_t$ changes. Variable $R_t$ may be encountered in applications like video surveillance, where the available bit rate on the communication channel may vary over time. The particular convex surface we use for approximating the measured rate-distortion points is:       
\begin{equation}
\label{eq:dist_total_form}
    D_t(R_1,...,R_N) \approx \gamma + \sum_{j=1}^{N} \alpha_j 2^{-\beta_j R_j}
\end{equation}
where 
$\alpha_j$, $\beta_j$, and $\gamma$ are surface parameters. 
In our experiments, we used non-linear least squares method based on Trust Region Reflective Algorithm \cite{scipy_fit} 
to fit the surface~(\ref{eq:dist_total_form}) to the rate-distortion points. As an example, Fig.~\ref{fig:surface} shows a fitted surface for a multi-task model used in our experiments, with two tensors to be coded (hence, two rates). As seen in the figure, the agreement between the original points and the fitted surface is quite good. This is further confirmed quantitatively using the coefficient of multiple determination $R^2$~\cite{Neter_etal_1988}, which, for the surface in Fig.~\ref{fig:surface}, was $R^2=0.99$. Note that $0\leq R^2 \leq 1$, so this value of $R^2$ is quite high. In addition, the residuals (the differences between the actual points and the fitted surface) were clustered around zero, with mean residual being $-1.4\times{10^{-8}}$. Together with the high value of $R^2$, this indicates that the model in~(\ref{eq:dist_total_form}) is an excellent approximation to the measured rate-distortion points. Indeed, in all test cases in our experiments we were obtaining $R^2>0.95$, 
with residuals centered around zero. 


Once the surface parameters are obtained, the solution to~(\ref{eq:rdo}) can be obtained using the standard method of Lagrange multipliers~\cite{Boyd_Vandenberghe_2004}. Specifically, the constrained problem in~(\ref{eq:rdo}) is converted to an unconstrained problem of minimizing the Lagrangian $J$, given by
\begin{equation}
    J = D_t(R_1,...,R_N) + \lambda \cdot \left(\sum_{j=1}^{N} R_j - R_t\right),
\end{equation}
where $\lambda$ is the Lagrange multiplier. By solving the system of $N+1$ equations, 
\begin{equation}
    \frac{\partial J}{\partial \lambda}=0 \quad \text{and} \quad \frac{\partial J}{\partial R_j}=0, \enspace j=1,2,...,N,
\end{equation}
in $N+1$ unknowns $(\lambda, R_1, ..., R_N)$, we obtain the solution 
\begin{equation}
\label{eq:solution}
    R_j^* = \frac{1}{{\beta_j}}\left(
    {\log_2(\alpha_j \beta_j)-\frac{\sum_{k=1}^{N}{\frac{1}{\beta_k}\log_2(\alpha_k \beta_k)} - R_t}{\sum_{k=1}^{N}{\frac{1}{\beta_k}}}}\right)
\end{equation}
for $j=1,2,...,N$. Two points are worth noting about the above solution. First, the task weights $w_i$ do not appear explicitly in the expression for $R_j^*$, because they are subsumed by the surface parameters. 
That is to say, a different set of task weights would lead to a different set of surface parameters, which would lead to a different set of rates in~(\ref{eq:solution}). A change in task weights does not require another measurement of rate-distortion points, because task-specific distortions in~(\ref{eq:dist_def}) would stay the same. Only the total distortion~(\ref{eq:total_dist_def}) would change, and this would require a new surface fitting in~(\ref{eq:dist_total_form}) to obtain the new surface parameters. The second point to note is that the solutions in~(\ref{eq:solution}) may be negative for some values of $j$. We have not encountered such cases in our experiments, but in practice, one would clip $R_j^*$ from below at $0$, which is a standard approach in bit allocation~\cite{Gersho_Gray_1992}.   

\section{Experiments}
\label{sec:Experiments}
To evaluate the proposed bit allocation method, we trained the model in Fig.~\ref{fig:model} using the Cityscapes dataset~\cite{cityscapes}. The three tasks are semantic segmentation, disparity estimation, and input reconstruction.  Cityscapes dataset includes 2,975 training images with their corresponding semantic segmentation and disparity maps. Since the annotations for the test set are not publicly available, the 500 images in the validation set are used as the test set, as in~\cite{cambridge, NeurIPS_Sener}. 

The backbone 
(part of the model between the input and the stack block in Fig.~\ref{fig:model}) is similar to the backbone of YOLOv3~\cite{YOLOv3} and has 74 convolutional layers. Its weights are initialized using YOLOv3 weights. Models 1, 2, and 3 on the cloud side are based on the FC8 model~\cite{FC} and use the stacked deep features as inputs. 
Cross-entropy loss~\cite{metrics} is used for semantic segmentation and Mean Square Error (MSE) is used as the loss function for the other two tasks. 
Following~\cite{cambridge}, the weighted sum of the mentioned losses is used as the total loss, and each task's weight is a trainable parameter which is trained during the training process.
Adam optimizer with the initial learning rate of 0.1 and rate decay by a factor of 0.85 every 20 epochs is used to train the three-task model, end to end, for 250 epochs. 

Once trained, two features tensors are taken from the model: one at layer 36 and the other from layer 61  (these features stream skip to the stack block), so 
$N=2$ in~(\ref{eq:rdo})--(\ref{eq:solution}). In this case, the total distortion $D_t$ is a function of two rates, so it can be displayed as a surface in 3D, as shown in Fig.~\ref{fig:surface}. 
We initially apply uniform 8-bit (min-max) quantization to the deep feature tensors, then rearrange their channels to form tiled images, as in~\cite{choi_icip}. 
Any image codec can be used to encode the tiled tensor image; 
for our experiments we used  JPEG2000~\cite{jpg2k} because of its rate control tools that allow us to obtain a desired rate fairly accurately. 

By choosing 100 rate pairs $(R_1^k,R_2^k)$, $k=1,2,...,100$, encoding the tiled tensor images obtained from 20\% of the samples in the training set at these rates, measuring the model's performance using decoded tensors and computing the total distortion $D_t^k$ from~(\ref{eq:total_dist_def}), we obtain 100 rate-distortion triplets $(R_0^k,R_1^k,D_t^k)$. A sample rate-distortion surface 
(with $w_i=1$ in~(\ref{eq:total_dist_def})) 
and its convex approximation~(\ref{eq:dist_total_form}) are shown in 
Fig.~\ref{fig:surface}. 
For the approximation surface shown in Fig.~\ref{fig:surface}, the obtained parameters (for the rates in Kbits) were $\alpha_1=72.45, \alpha_{2}=183.09, \beta_1=7.07 \times10^{-4}, \beta_{2}=2.11 \times 10^{-2}, \gamma=0.80$. If a different set of task weights $w_i$ is desired, we simply recompute the total distortion $D_t^k$ from~(\ref{eq:total_dist_def}); task-specific distortions $D_i^k$ in~(\ref{eq:dist_def}) do not change, unless the rates $(R_1^k, R_2^k)$ change. 



\begin{table}[t]
\vspace{-0.3cm}
\centering
\caption{Total distortion $D_t$, with $w_i=1$, achieved by various bit allocation methods for three rate constraints $R_t$ in kbits/tensor.}
\vspace{0.2cm}
\label{tbl:benchmarks_dist_comparison}
\begin{tabular}{|c|c|c|c|c|}
\hline
$R_t$ & Method 1 & Method 2 & Method 3 & Proposed  \\ \hline \hline  

$1000$ & 57.69 & 53.94   & 53.84   &  \textbf{53.62}  \\ \hline
$1500$& 50.97            &  46.57  & 46.49   & \textbf{43.32}         \\ \hline
$2000$& 46.16            & 38.96   & 38.77   &  \textbf{34.02}        \\ \hline
\end{tabular}
\end{table}

\begin{figure}[t]	
	\centering
	\centerline{\includegraphics[scale=0.52]{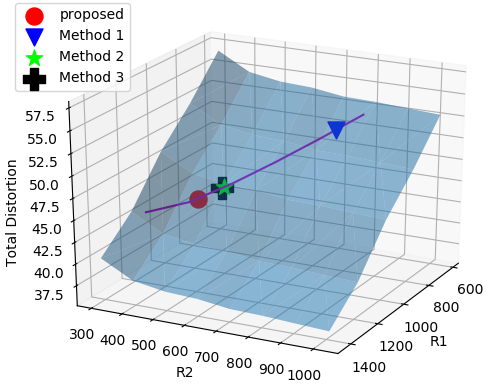}}
	\caption{Distortions achieved by various methods for $R_t=1500$. Rate-distortion surface is transparent blue, and the 
	magenta line is the set of points for which $R_1 + R_2=1500$.
	} 
	\label{fig:surface_comparison}
\end{figure}

After the parameters $(\alpha_j,\beta_j,\gamma)$ are obtained, for a given total rate $R_t$ 
we use~(\ref{eq:solution}) to obtain the optimal $R_j^*$, $j=1,2$. 
This solution is tested on the test set, and compared against three alternatives: equal rate allocation (Method 1), rate allocation proportional to the number of tensor elements (Method 2), and rate allocation proportional to the variance of tensor elements (Method 3). 
Table~\ref{tbl:benchmarks_dist_comparison} shows the total distortion $D_t$ with $w_i=1$  in~(\ref{eq:total_dist_def}) for three rate constraints: $R_t\in\{1000, 1500, 2000\}$ kbits/tensor. The lowest distortion under each $R_t$ is indicated in bold. At the lowest of these rates ($R_t=1000$), the rate allocation computed from~(\ref{eq:solution}) is relatively close to the ones obtained by Methods 2 and 3, and the resulting distortions are also similar. Nonetheless, the proposed solution gives the lowest distortion. The gap between the distortion achieved by the proposed bit allocation method and those obtained by alternative methods increases as $R_t$ increases. Fig.~\ref{fig:surface_comparison} illustrates distortions achieved by various methods for $R_t=1500$.  


Next we examine the effects of task weights $w_i$ in~(\ref{eq:total_dist_def}). When the weights change, the total distortion $D_t$ and its approximating surface in~(\ref{eq:dist_total_form}) will change, so the proposed method will find different bit allocations in~(\ref{eq:solution}). Meanwhile, the three benchmark methods keep their bit allocations, because the number of elements in the tensors and their variance stay the same. Their task-specific accuracies also stay the same, but their total distortion changes according to the new weights~(\ref{eq:total_dist_def}).
Table~\ref{tbl:benchmarks_result_comparison} shows both the total distortions (bottom two rows) and task-specific accuracy (middle three rows) for two weight settings: $(w_1,w_2,w_3)=(1,1,1)$ and $(w_1,w_2,w_3)=(8,1,1)$, when $R_t=1500$. The accuracy of semantic segmentation is measured by mean Intersection over Union (mIoU)~\cite{metrics}, the accuracy of disparity estimation is measured by Root Mean Squared (RMS) error in pixels~\cite{metrics}, and the quality of input reconstruction is measured in Peak Signal to Noise Ratio (PSNR) in dB.  

\begin{table}[b!]
\caption{Task-specific accuracies and total distortion  for $R_t$=1500. Higher numbers are better for mIoU and PSNR; lower numbers are better for RMS.}
\vspace{0.2cm}
\label{tbl:benchmarks_result_comparison}
\resizebox{\columnwidth}{!}{%
\begin{tabular}{|c|c|c|c|c|c|}
\hline
$R_t=1500$                                                             & \begin{tabular}[c]{@{}c@{}}Method\\ 1\end{tabular} & \begin{tabular}[c]{@{}c@{}}Method \\ 2\end{tabular} & \begin{tabular}[c]{@{}c@{}}Method \\ 3\end{tabular} & \begin{tabular}[c]{@{}c@{}}Prop.\\ (1, 1, 1)\end{tabular} & \begin{tabular}[c]{@{}c@{}}Prop.\\ (8, 1, 1)\end{tabular} \\ \hline \hline 
mIoU (\%)                                                              & 62.59                                              & 62.67                                               & 62.68                                               & 62.43                                                  & 62.66                                                     \\ \hline
RMS (px)                                                               & 7.85                                               & 7.86                                                & 7.86                                                & 7.88                                                   & 7.85                                                      \\ \hline
PSNR (dB)                                                              & 22.40                                              & 24.33                                               & 24.35                                               & 26.07                                                 & 24.90                                                     \\ \hline \hline 
\begin{tabular}[c]{@{}c@{}}
$D_t$\\ (1, 1, 1)\end{tabular}    & 50.97                                              & 46.57                                               & 46.49                                               & \textbf{43.32}                                         & N/A                                                        \\ \hline
\begin{tabular}[c]{@{}c@{}}
$D_t$\\ (8, 1, 1)\end{tabular} & 55.31                                             & 50.06                                              & 49.80                                              & N/A                                                     & \textbf{48.88}                                                     \\ \hline
\end{tabular}%
}
\end{table} 

Table~\ref{tbl:benchmarks_result_comparison} shows that when  $(w_1,w_2,w_3)=(1,1,1)$, the proposed method achieves much better (by at least 1.7 dB) input reconstruction than the other methods, but slightly worse disparity estimation (0.03px higher than the best result) and semantic segmentation (0.25\% lower than the best result). The total distortion of the proposed method is, of course, better than the other methods (fourth row), but with the weights $(w_1,w_2,w_3)=(1,1,1)$, the proposed rate allocation makes one task accuracy much better, and the other two slightly worse, compared to the next best result. 
On the other hand, by increasing the weights of the first task (semantic segmentation) to 8,  $(w_1,w_2,w_3)=(8,1,1)$, we can achieve the results as accurate as the best result among the benchmark methods on these two tasks, while still outperforming all of the benchmarks on input reconstruction, as well as the total distortion. 
This shows that the weights can be used to achieve a desired balance of task accuracies, depending on the application. 

\section{Conclusion}
\label{sec:conclusion}
In this paper we introduced the bit allocation problem for multi-stream, multi-task collaborative intelligence. A convex approximation to the rate-distortion surface was proposed, which led to the closed-form solution. 
The experiments showed that the proposed bit allocation  results in lower total distortion for the given rate constraint compared to several alternative bit allocation methods. 

\bibliographystyle{IEEEbib}
\bibliography{refs_cr}

\end{document}